\documentclass{article}
\usepackage[preprint]{spconf}
\usepackage{amsmath}
\usepackage{graphicx}
\usepackage{booktabs}
\usepackage{multirow}
\usepackage{amssymb}
\usepackage{xcolor}
\usepackage{hyperref}

\usepackage{amsmath} 
\usepackage{float}

\newcommand\blfootnote[1]{%
  \begingroup
  \renewcommand\thefootnote{}\footnote{#1}%
  \addtocounter{footnote}{-1}%
  \endgroup
}

\title{Towards Building Text-To-Speech Systems for the Next Billion Users}
\name{%
\begin{tabular}{@{}c@{}}
Gokul Karthik Kumar\sthanks{Equal contribution}\sthanks{Work started with internship at Microsoft and concluded at MBZUAI}$^{1,3,4}$ \qquad
Praveen S V $^{*1,2}$\\
Pratyush Kumar$^{1,2,4}$ \qquad 
Mitesh M. Khapra$^{1,2}$ \qquad 
Karthik Nandakumar$^3$
\end{tabular}}
\address{$^1$AI4Bharat, India \\ $^2$Indian Institute of Technology Madras (IITM), India \\ $^3$Mohamed Bin Zayed University of Artificial Intelligence (MBZUAI), UAE \\ $^4$Microsoft Research, India }

\begin{document}
\ninept
\maketitle

\begin{abstract}
Deep learning based text-to-speech (TTS) systems have been evolving rapidly with advances in model architectures, training methodologies, and generalization across speakers and languages.
However, these advances have not been thoroughly investigated for Indian language speech synthesis.
Such investigation is computationally expensive given the number and diversity of Indian languages, relatively lower resource availability, and the diverse set of advances in neural TTS that remain untested. 
In this paper, we evaluate the choice of acoustic models, vocoders, supplementary loss functions, training schedules, and speaker and language diversity for Dravidian and Indo-Aryan languages.
Based on this, we identify monolingual models with FastPitch and HiFi-GAN V1, trained jointly on male and female speakers to perform the best.
With this setup, we train and evaluate TTS models for 13 languages and find our models to significantly improve upon existing models in all languages as measured by mean opinion scores.
We open-source all models on the Bhashini platform \footnote{\scriptsize \url{https://bhashini.gov.in/ulca/model/explore-models}}.
\blfootnote{Copyright 2023 IEEE. Published in ICASSP 2023 – 2023 IEEE International Conference on Acoustics, Speech and Signal Processing (ICASSP), scheduled for 4-9 June 2023 in Rhodes Island, Greece. Personal use of this material is permitted. However, permission to reprint/republish this material for advertising or promotional purposes or for creating new collective works for resale or redistribution to servers or lists, or to reuse any copyrighted component of this work in other works, must be obtained from the IEEE. Contact: Manager, Copyrights and Permissions / IEEE Service Center / 445 Hoes Lane / P.O. Box 1331 / Piscataway, NJ 08855-1331, USA. Telephone: + Intl. 908-562-3966.}


\end{abstract}
%

\section{Introduction}
\label{sec:intro}


Deep neural networks have led to rapid progress in text-to-speech (TTS) systems. Compared to traditional methods like formant, concatenative, and statistical parametric speech synthesis, neural TTS achieves high-fidelity real-time speech synthesis with limited need for manual feature engineering \cite{tan2021survey}. This has enabled the generation of high-quality synthetic speech, which is being increasingly used in a larger number of applications.

A TTS system consists of 3 principal components: a text analysis module that converts text to linguistic features, an acoustic model that converts linguistic features to acoustic features, and a vocoder that converts acoustic features to speech waveforms. 
Many of the recent state-of-the-art TTS systems use two-stage speech synthesis models \cite{shen2018natural, kim2020glow, ren2019fastspeech, lancucki2021fastpitch}, which amalgamate the first two components with an acoustic model to directly convert text to features such as spectrograms, or omit the text analysis module entirely.
Within this class of models, several advances have been made.
WaveNet \cite{oord2016wavenet} was one of the earliest works based on recurrent neural networks (RNNs) to generate speech waveforms directly from linguistic features.
Tacotron \cite{wang2017tacotron} was the first successful neural acoustic model to generate spectrograms from the text directly.
The use of an autoregressive TTS based on RNNs to generate speech waveforms was demonstrated in Tacotron2 \cite{shen2018natural}.
The speed of the acoustic models was improved by replacing RNNs with Transformer-based non-autoregressive (NAR) acoustic models as demonstrated in FastSpeech \cite{ren2019fastspeech} and FastPitch \cite{lancucki2021fastpitch}.
However, these NAR models require an external aligner module.
The need for this aligner module was eliminated with the proposal of flow-based generative models, such as Glow-TTS \cite{kim2020glow}, which implements a monotonic alignment search algorithm to map latent speech representations to representations in the text domain. 
Meanwhile, several neural vocoders like WaveGAN \cite{donahue2018adversarial}, MelGAN \cite{kumar2019melgan}, HiFiGAN \cite{kong2020hifi} and Multi-Band MelGAN \cite{yang2021multi} adapted Generative Adversarial Networks (GANs) for generating audio waveforms, which improve quality of generated speech, primarily with changes in the discriminator and addition of new loss functions.
Recently, neural speech synthesizers \cite{jeong2021diff,popov2021grad} based on denoising probabilistic diffusion have been proposed which generate high-quality speech but tend to be slower in inference owing to their iterative nature. 
While two-stage TTS systems remain popular, there is an ongoing exploration of end-to-end systems, such as VITS \cite{kim2021conditional}, that  directly synthesizes speech from text.

Apart from advances in neural architectures, there has been an interest in developing TTS systems for low-resource settings. 
One approach is to study multi-speaker generalization. 
This has been studied for English \cite{kim2020glow, lancucki2021fastpitch,  kim2021conditional}, with models that can generate speech for multiple speakers as represented by speaker embeddings.
Such models also have the practical benefit of efficient deployment in supporting multiple voices (say, one male and one female)  from a single hosted model.
Another approach is to consider multilingual generalization 
\cite{chen2019end} to transfer knowledge from high resource languages by mapping the embeddings of the phoneme sets from different languages. 
Recently, YourTTS \cite{casanova2022yourtts} successfully extended the end-to-end VITS model for multilingual generalization by conditioning on language embeddings. 

The above paragraphs briefly summarize the advances in neural TTS over half a decade of active research.
A characteristic of TTS, somewhat different from other domains such as computer vision, language modelling, neural translation, and speech recognition, is that there is a large diversity of neural architectures and modelling techniques that continue to remain competitive.
In other words, there is no one dominant TTS design methodology that is conclusively superior.
Thus, the task of bringing the latest advances in TTS research to a set of languages requires that various design methodologies be implemented and tested with human evaluation. 
This is particularly challenging for Indian languages which are not only numerous but also significantly differ in terms of phonetics, morphology, word semantics, syntax and written scripts.

There have been a few studies specifically focused on Indian languages.
For example, Vakyansh \cite{vakyansh2021glow} open-sourced TTS models for 9 Indian languages with a combination of Glow-TTS with HiFi-GAN.
Similarly, multilingual TTS models for Indian languages within the same family have been built \cite{prakash2020generic} using Tacotron2  with WaveGlow \cite{prenger2019waveglow} by making use of the multi-lingual character map \cite{prakash2019building} and the common label set \cite{baby2016unified}. 
However, various recent advances in TTS systems remain to be tested for Indian languages.
For instance, TTS systems have not been built and evaluated for Indian languages that exhibit the following: fast generation as in FastPitch model \cite{lancucki2021fastpitch}, flow-based generation as in GlowTTS model \cite{kim2020glow}, end-to-end generation as in VITS model \cite{kim2021conditional}, comparison of GAN based vocoders as in HiFiGAN \cite{kong2020hifi} and Multi-Band MelGAN\cite{yang2021multi}  with conditional models for waveform generation as in WaveGrad \cite{chen2020wavegrad}. 
Also, the efficacy of multi-speaker models for Indian languages remains untested. 
Finally, multi-lingual models that group languages by their language family need to be explored for the latest state-of-the-art models.

In this paper, we aim to partially address this gap with a rigorous exploration of various TTS systems for 13 Indian languages across choices of acoustic models, vocoders, supplementary loss functions, training schedules, and speaker and language generation.
Specifically, we consider three different acoustic models - FastPitch \cite{lancucki2021fastpitch}, GlowTTS \cite{kim2020glow}, and the end-to-end VITS \cite{kim2021conditional}, three different vocoders - HiFiGAN V1 \cite{kong2020hifi}, Multi-Band MelGAN \cite{yang2021multi}, and WaveGrad \cite{chen2020wavegrad}, use of SSIM \cite{wang2004image} based supplementary loss, multi-speaker training with male and female voices, and multilingual models for four Dravidian, seven Indo-Aryan, and two Sino-Tibetan languages.
With manual MOS-based and automated metric-based evaluation, we identify the combination of FastPitch and HiFiGAN V1, trained with male and female speakers, for a single language to be the preferred setup.
With this setup, we train TTS models for the 13 languages and establish that these models improve upon existing TTS models with both MOS and automated metrics.

The real-world impact of natural sounding TTS is significant in a country like India with the need to deliver digital touch-points to a large population speaking 122 major languages across 5 different language families, with 25\% of the population with reading disabilities. 
Thus, there is value for ``foundation models'' \cite{bommasani2021opportunities} in TTS to be available in the open-source to enable rapid innovation and deployment.
This movement towards open-sourcing has been gaining momentum in other areas such as language modelling \cite{devlin2018bert, kakwani2020indicnlpsuite}, transliteration \cite{madhani2022aksharantar}, translation \cite{costa2022no}, and speech recognition \cite{javed2022towards}.
To contribute towards this effort, we open-source all our TTS models through the Bhashini platform \cite{bhashini}.

\section{Design Choices for TTS Models}
\label{sec:design-choices}

In this section, we detail the different model architectures, training strategies, and generalization techniques that we evaluate for TTS models for Indian languages.
\subsection{Acoustic models}
\label{subsec:acoustic}

\textbf{FastPitch}\cite{lancucki2021fastpitch}: 
To represent fast NAR models and acoustic models based on Transformers, we consider FastPitch \cite{lancucki2021fastpitch}.
The model is based on the feed-forward Transformer consisting of an encoder, 1-D convolution based duration and pitch predictors, and a decoder. 
The pitch and duration predictors are trained using mean-squared-error losses, but unlike \cite{lancucki2021fastpitch} we extract ground truth frequencies from WORLD \cite{morise2016world} and train the duration predictor on durations learnt from an alignment learning framework \cite{badlani2022one}.\\
\textbf{Glow-TTS} \cite{kim2020glow}: 
To represent flow-based generative models we consider Glow-TTS. Similar to FastPitch, Glow-TTS uses a Transformer based encoder with slight modifications \cite{kim2020glow}. The model eliminates the need for an external aligner and uses the default Monotonic Alignment Search (MAS) algorithm trained with maximum likelihood estimation. The decoder is composed of a family of invertible flows and transforms a prior distribution into mel-spectrograms.\\
\textbf{VITS} \cite{kim2021conditional}: To represent end-to-end models, we consider VITS \cite{kim2020glow}. The model uses the same text-encoder as Glow-TTS \cite{kim2020glow}, a posterior encoder consisting of non-causal residual WaveNet blocks, a decoder based on HiFiGAN-V1 \cite{kong2020hifi} a multi-period discriminator and a stochastic duration predictor.

We did not consider autoregressive methods such as Tacotron2 as Glow-TTS, Fast Pitch and VITS have been demonstrated to be better. However, as a limitation of our work, we do not include the most recent diffusion-based acoustic models such as Grad-TTS \cite{popov2021grad}.

\subsection{Vocoders}
\label{subsec:vocoders}
For the choice of vocoders, we restrict ourselves to those that use mel-spectrograms as the input representation. 
We do not consider auto-regressive models such as WaveNet \cite{oord2016wavenet} which have been improved upon by other vocoders both on generation speed and quality. In this work, we consider the following vocoders - \\
\textbf{HiFi-GAN V1} \cite{kong2020hifi}: The neural vocoder is based on a generative adversarial network and achieves high computational efficiency and sample quality. By using a single generator and multi-scale and multi-period discriminators, operating on different scales and periods of the input waveform, it captures implicit structures and long-term dependencies and is able to efficiently generate high-fidelity speech. \\
\textbf{Multi-Band MelGAN} \cite{yang2021multi}: This neural vocoder extends MelGAN \cite{kumar2019melgan} by doubling the receptive field for improved speech generation. It also substitutes the feature matching loss in MelGAN \cite{kumar2019melgan} with a multi-resolution STFT loss evaluated at both sub-band and full-band scales. Further, by adopting shared parameters for all sub-band signal predictions, Multi-Band MelGAN is able to achieve better speech with faster generation speeds.\\
\textbf{WaveGrad} \cite{chen2020wavegrad}: We also consider a diffusion-based vocoder, namely WaveGrad \cite{chen2020wavegrad}, which is a non-autoregressive neural vocoder that iteratively transforms white Gaussian noise into high-quality audio waveforms by using a gradient-based sampler conditioned on mel-spectrograms. \\
As a limitation of our work, we do not consider flow-based vocoders such as WaveGlow. \cite{prenger2019waveglow}.


\subsection{Training Strategies}
\label{subsec:training-strategies}
Different TTS works employ different loss functions to enhance training, such as revisiting SSIM loss \cite{ren2022revisiting} or employing an ASR-based speech consistency loss \cite{li2021starganv2} to enhance training.
In this work, we consider two supplementary loss functions - the SSIM loss on the synthesized mel-spectrogram and a speech consistency loss \cite{wang2004image} that characterizes the intelligibility of the generated speech.
The SSIM loss measures the structural similarity between the synthesized mel-spectrogram and ground truth mel-spectrogram on three dimensions: luminance, contrast and structure.
The ASR loss measures the $l_1$-norm between convolutional features of the ground truth and synthesized mel-spectrograms extracted from the intermediate layers of the pretrained joint CTC-attention VGG-BLSTM network \cite{li2021starganv2}.
In addition, we consider training schedules where the alignment loss is turned off after a fixed number of steps. 


\subsection{Multi-speaker models}
When building multi-speaker models, a common approach is to introduce speaker embeddings to condition the acoustic models.
There are two ways of computing speaker embeddings: learn the speaker embedding network while training the acoustic model \cite{gibiansky2017deep, ping2018deep} and use an external pretrained speaker verification model \cite{wan2018generalized} to pre-compute embeddings.
Since pretrained speaker verification models for Indian languages are not easily available, we use the former approach. 
We learn speaker embeddings and perform a point-wise vector addition with the encoder output of the acoustic model.  



\subsection{Multilingual models}
To train multilingual models, we condition the text encoder outputs on language IDs using learnable embeddings optimized during training. 
All our acoustic models take in the raw text as input, instead of phonemes.
To aid the generalization of the representations computed by the text encoder, we choose to map the diverse scripts of different Indian languages into a common representation. 
We do this by transliterating all scripts to the ISO format with the help of Aksharamukha \footnote{ \scriptsize\url{https://aksharamukha.appspot.com/}}. As future work, we would also like to explore the common label set for Indian languages proposed in \cite{baby2016unified}.

\section{Experiments}

\subsection{Experimental setup}

\begin{figure}[htp]
    \centering
    \includegraphics[width=0.95\columnwidth]{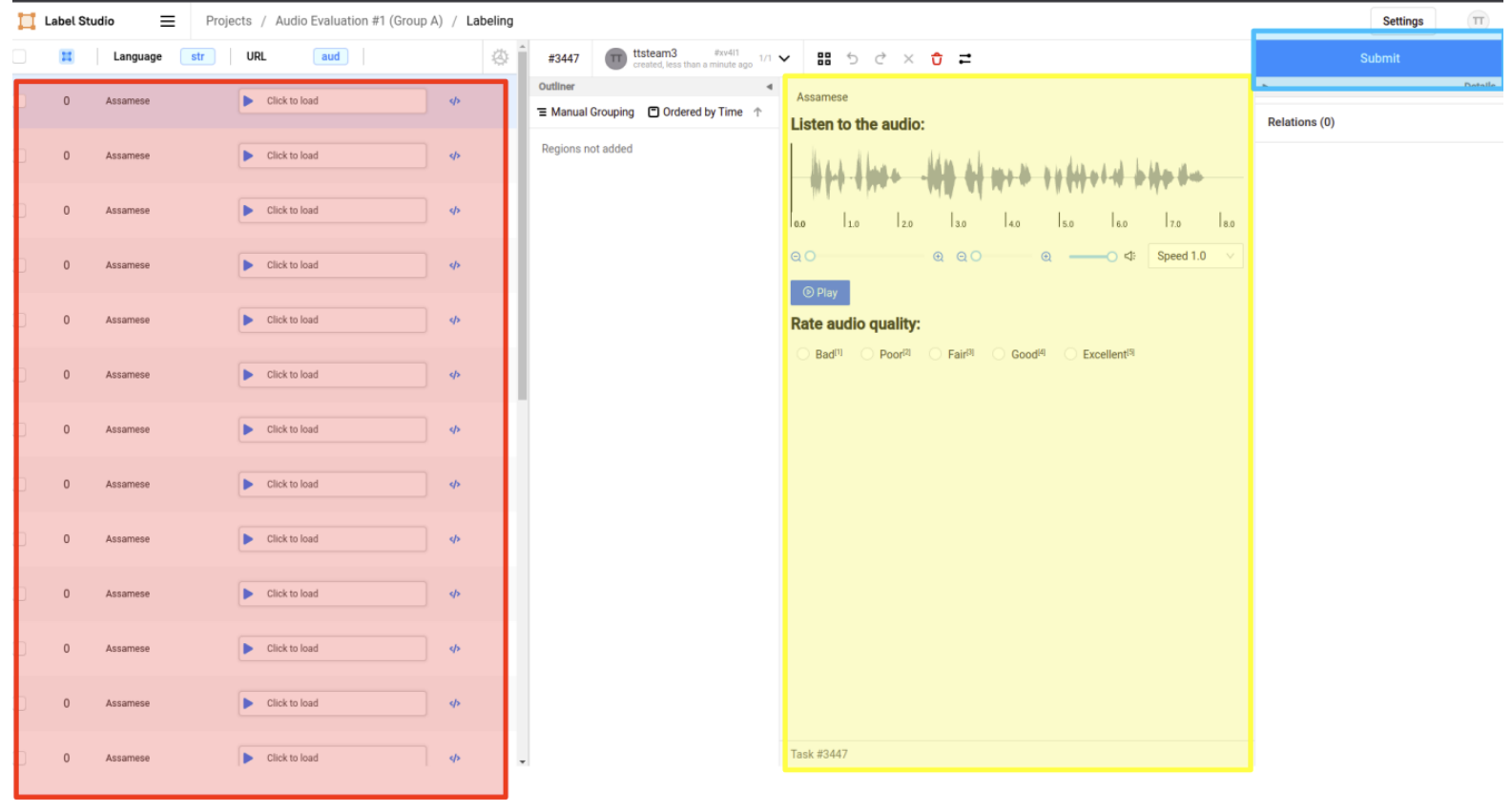}
    \caption{Annotation Interface for MOS Evaluation on Label Studio. Raters use the red-shaded pane to navigate across audio samples, the yellow-shaded pane to rate the sample and submit their final ratings using the blue submit button.}
    \label{fig:labelstudio}
\end{figure}

\textbf{Dataset:} We use the latest version of the IndicTTS Database \cite{baby2016resources} with over 272 hours of transcribed speech recordings for 13 Indian languages including Assamese, Bengali, Bodo, Gujarati, Hindi, Kannada, Malayalam, Manipuri, Marathi, Odia, Rajasthani, Tamil and Telugu. All languages except Bodo have both male and female speakers, while Bodo only has a female speaker. For each speaker, there exists at least 8 hours of transcribed data. All audio samples are downsampled to a sampling rate of 22.05KHz and utterances having a duration greater than 20 seconds are filtered out. We preprocess text by replacing semi-colons and colons with commas, removing parenthesis symbols and collapsing whitespaces.
\\
\textbf{Training \& Inference:} We implement our models using Coqui-TTS\footnote{\scriptsize \url{https://github.com/coqui-ai/TTS}} library. All models are trained for 2500 epochs on a single NVIDIA A100 40GB Tensor Core GPU, with a batch size of 32 and the default learning rate scheduling of Coqui-TTS. We train FastPitch with Adam optimizer with $\beta_1=0.99$ and $\beta_2=0.998$ with weight decay of $\lambda=10^{-6}$. Glow-TTS is trained with RAdam optimizer with $\beta_1=0.99$ and $\beta_2=0.998$ with weight decay of $\lambda=10^{-6}$. VITS is trained using the AdamW optimizer with $\beta_1=0.8$ and $\beta_2=0.99$ with weight decay of $\lambda=0.01$. Each model took approximately 3 days to train. For our final models in Section \ref{subsec:best}, we turned off the aligner for the last 1000 epochs, as this helped us achieve better convergence in spectrogram reconstruction. While training vocoders, we use the default hyper-parameter settings in Coqui-TTS, except for WaveGrad where we increase the batch size to 96. We train a separate vocoder for each language. We observed that having large variations in average utterance durations between individual speakers for a given language make it difficult for multi-speaker acoustic models to learn alignments between input text and mel-spectrograms. 
For example, while training a multi-speaker model for the Telugu speaker, where the average utterance duration for the female speaker was nearly twice that of the male speaker, we observed the alignment loss failed to converge. This was resolved by modulating the tempo of the Telugu female's utterances to be 0.77x its original speed. 
Multilingual models for each language family are trained with the help of the Aksharamukha tool that supports transliteration for 10 of 13 Indian languages excluding Bodo, Manipuri, and Rajasthani. 
We post-process the generated audio samples with DCCRN \cite{hu2020dccrn} speech enhancement model to remove background artefacts.
\\
\textbf{Evaluation:}  We evaluate our models using subjective and objective metrics on a validation set of 30 utterances unseen during training. We conduct a subjective Mean Opinion Score (MOS) evaluation on LabelStudio \cite{label-studio} with the help of 42 raters, all of whom are native speakers of the language they are tasked to evaluate. This includes 6 raters each for Tamil and Hindi, 1 rater for Rajasthani and 3 raters for each of the remaining 10 languages. In Figure \ref{fig:labelstudio}, we depict the annotation interface, where evaluators are requested to rate each audio sample on a scale of five [1 - Bad, 2 - Poor, 3 - Fair, 4 - Good and 5 - Excellent]. 


To measure acoustics objectively, we use two metrics: mel-cepstral distortion (MCD) \cite{kubichek1993mel} and root-mean-square error of the log of the fundamental frequencies ($F_0$), with dynamic time warping \cite{salvador2007toward} to temporally align the sequences. To measure intelligibility objectively, we use the character error rate (CER) with text extracted from Google Cloud's Automatic Speech Recognition\footnote{\scriptsize \url{https://cloud.google.com/speech-to-text}}. Dravidian and Indo-Aryan languages have very distinct characteristics \cite{prakash2020generic}.
As it is computationally expensive to experiment with all 13 Indian languages, we choose one language under each family: Tamil (Dravidian) and Hindi (Indo-Aryan) to evaluate the design choices for TTS models. 


\subsection{Evaluation of acoustic models and neural vocoders} \label{subsec:combinations}

\begin{table*}[ht]
\centering
\begingroup
\setlength{\tabcolsep}{6pt} 
\renewcommand{\arraystretch}{0.8} 
\begin{tabular}{@{}lllllrrrccccccc@{}}
\toprule
\multicolumn{1}{c}{Metrics} & \multicolumn{4}{c}{MOS}                               & \multicolumn{3}{c}{MCD}                                    & \multicolumn{3}{c}{$F_0$}                                    & \multicolumn{4}{c}{CER}             \\ 
\cmidrule(l){1-1} \cmidrule(l){2-5} \cmidrule(l){6-8} \cmidrule(l){9-11} \cmidrule(l){12-14} 
\textbf{Language}                    & \textbf{GT} & \textbf{Ours} & \textbf{D} & \textbf{V} & \textbf{Ours}  & \textbf{D} & \textbf{V}                            & \textbf{Ours} & \textbf{D} & \textbf{V}                           & \textbf{GT} & \textbf{Ours}  & \textbf{D} & \textbf{V}     \\
\midrule
\multicolumn{14}{c}{Dravidian Languages}                           \\
\midrule
Kannada    & 4.11 & \textbf{3.68} & 2.90 & 2.27 & \textbf{8.47}  & 13.11 & 9.64          & \textbf{0.27} & 0.35 & 0.30         & 0.132 & \textbf{0.120} & 0.217 & 0.391          \\
Malayalam  & 4.24 & \textbf{3.64} & 3.49 & 1.92 & \textbf{8.08}  & 12.48 & 9.31          & \textbf{0.18} & 0.21 & 0.22         & 0.123 & \textbf{0.216} & 0.303 & 0.337          \\
Tamil      & 4.16 & \textbf{3.84} & 3.01 & 2.59 & 10.57          & 11.92 & \textbf{7.28} & \textbf{0.31} & 0.38 & 0.35         & 0.108 & \textbf{0.107} & 0.245 & 0.134          \\
Telugu     & 4.42 & \textbf{3.66} & 3.40 & 2.61 & 9.91           & 11.03 & \textbf{7.16} & \textbf{0.36} & 0.38 & 0.37         & 0.273 & \textbf{0.266} & 2.254 & 0.364          \\
\midrule
\multicolumn{14}{c}{Indo-Aryan Languages}                                                                           \\ 
\midrule
Assamese   & 3.63 & \textbf{2.39} & -    & -    & \textbf{10.93} & -     & -             & \textbf{0.38} & -    & -            & - & -              & -     & -              \\
Bengali    & 4.58 & \textbf{3.37} & -    & 3.16 & 10.18          & -     & \textbf{5.97} & \textbf{0.28} & -    & \textbf{0.28} & 0.145 & \textbf{0.167} & -     & 0.193          \\
Gujarati   & 4.12 & \textbf{3.58} & -    & 3.02 & 8.99           & -     & \textbf{5.56} & 0.38          & -    & \textbf{0.36} & 0.156 & 0.194          & -     & \textbf{0.182} \\
Hindi      & 4.33 & \textbf{4.00} & 3.70 & 3.16 & 8.73           & 12.41 & \textbf{8.50} & \textbf{0.21} & 0.23 & 0.24        & 0.104  & \textbf{0.094} & 0.127 & 0.100          \\
Marathi    & 4.30 & \textbf{3.26} & 2.69 & 2.61 & \textbf{9.51}  & 14.84 & 11.53         & \textbf{0.30} & 0.41 & 0.50         & 0.093 & \textbf{0.075} & 0.136 & 1.484          \\
Odia       & 4.77 & \textbf{4.19} & 3.04 & 3.56 & \textbf{8.90}  & 15.72 & 11.07         & \textbf{0.22} & 0.33 & 0.24         & - & -              & -     & -              \\
Rajasthani & 4.10 & \textbf{3.40} & 3.37 & -    & \textbf{9.72}  & 12.89 & -             & \textbf{0.25} & 0.30 & -             & - & -              & -     & -               \\
\midrule
\multicolumn{14}{c}{Sino-Tibetan Languages}                                                                           \\ 
\midrule
Bodo       & 4.53 & \textbf{3.53} & -    & -    & \textbf{9.89}  & -     & -             & \textbf{0.23} & -    & -            & - & -              & -     & -              \\
Manipuri   & 4.58 & \textbf{3.30} & -    & -    & \textbf{11.68} & -     & -             & \textbf{0.27} & -    & -            & - & -              & -     & -              \\
\bottomrule
\end{tabular}
\endgroup
\caption{Results of our model and existing works on the IndicTTS Database in terms of acoustic metrics (MCD, $F_0$), intelligibility (CER) and subjective scores (MOS) for evaluating naturalness of generated samples. GT: Ground Truth, Ours: AI4Bharat-TTS FastPitch+HiFiGAN, D: DON Lab's Tacotron2+WaveGlow \cite{prakash2020generic}, V: Vakyansh's GlowTTS+HiFiGAN\cite{vakyansh2021glow}}
\label{tab:main}
\end{table*}

\begin{table}[htp]
\centering
\begingroup
\setlength{\tabcolsep}{2pt} 
\renewcommand{\arraystretch}{1} 
\begin{tabular}{@{}llcclccl@{}}
\toprule
\multicolumn{1}{c}{\textbf{Model}} & \multicolumn{1}{c}{\textbf{Vocoder}} & \multicolumn{3}{c}{\textbf{Tamil }}                                                & \multicolumn{3}{c}{\textbf{Hindi}}       \\ 
\multicolumn{1}{c}{} & \multicolumn{1}{c}{} & \multicolumn{3}{c}{(Dravidian)}                                                & \multicolumn{3}{c}{(Indo-Aryan)} \\ 
 \cmidrule(l){3-5} \cmidrule(l){6-8}
                                   &                                      & MCD        & \emph{$F_0$}          & \multicolumn{1}{c}{CER} & MCD        & $\mathbf{F_0}$          & \multicolumn{1}{c}{CER} \\
\midrule                                   
FastPitch  & HiFiGAN V1              & 11.19 & 0.30 & 0.103 & 7.59  & 0.21 & 0.095 \\
           & MB MelGAN      & 12.00 & 0.32 & 0.135 & 7.79  & 0.24 & 0.105 \\
           & WaveGrad              & 16.00 & 0.30 & 0.114 & 9.74  & 0.20 & 0.106 \\ \midrule
Glow-TTS   & HiFiGAN V1              & 11.73 & 0.33 & 0.204 & 8.36  & 0.24 & 0.198 \\
           & MB MelGAN      & 12.00 & 0.32 & 0.135 & 8.44  & 0.27 & 0.216\\
           & WaveGrad              & 16.90 & 0.31 & 0.243 & 10.30 & 0.24 & 0.191 \\ \midrule
VITS & \multicolumn{1}{c}{-} & 10.87 & 0.37 & 0.295 & 7.32  & 0.26 & 0.176\\  \bottomrule
\end{tabular}
\endgroup
\caption{Objective evaluation of a multi-speaker TTS system for different combinations of acoustic models and vocoders for Tamil and Hindi. Here, MB MelGAN refers to Multi-Band MelGAN}
\label{tab:combo}
\end{table}
We evaluate the combinations across acoustic models and vocoders for two languages objectively. In Table \ref{tab:combo}, we report objective metrics of combining the different acoustic models and vocoders mentioned in Sections \ref{subsec:acoustic} and \ref{subsec:vocoders} respectively. Within acoustic models and across languages, we observe a general trend of HiFi-GAN performing better in terms of acoustic metrics with a few exceptions where WaveGrad achieves slightly lower $F_0$ scores. Further, for each vocoder, FastPitch consistently outperforms Glow-TTS across all three metrics for both the languages. We also observe that VITS achieves the lowest MCD scores, potentially due to it being a fully end-to-end synthesizer. However, in comparison to FastPitch, the VITS model produces less intelligible speech as reflected with larger CER values and with average prosody as given by $F_0$ scores. Therefore, we pick the combination of FastPitch and HiFiGAN V1 as our model architecture  to build TTS systems for Indian languages. The chosen unified architecture  is elaborated in Appendix \ref{appendix:unified}.

\subsection{Evaluation of training strategies}
\label{subsec:loss}
As discussed earlier, we evaluate the use of supplementary SSIM and ASR loss functions.
When using the SSIM loss function, we observed a delayed convergence in the mel-spectrogram reconstruction loss.
However, the delay is not significant and both variations converge to similar values. 
Given no significant advantage, we choose to exclude the additional SSIM loss function.
We include the additional ASR loss and obtain intelligibility metrics for the two languages as shown in Table~\ref{tab:asr}. 
Since there are no significant improvements, we choose to exclude the ASR loss.
Thus, we do not add any supplementary loss functions to our training setup.

\begin{table}[!htbp]
\centering
\begingroup
\setlength{\tabcolsep}{6pt} 
\renewcommand{\arraystretch}{0.9} 
\begin{tabular}{@{}lcc@{}}
\toprule
\textbf{Language}  & {\textbf{Without ASR Loss}} & {\textbf{With ASR Loss}} \\ 
\midrule
Tamil (Dravidian)  & 0.107              & 0.108 \\ 
Hindi (Indo-Aryan) & 0.094             & 0.090  \\
\bottomrule
\end{tabular}
\endgroup
\caption{Objective evaluation on Intelligibility (CER) of our multi-speaker model with and without ASR loss for Tamil and Hindi.}
\label{tab:asr}
\end{table}

During our experiments, we observed that the alignment loss sharply rises at different points of the training, and subsequently all other losses would rise in response to this.
To address this, we experimented with a training schedule where we turned off the aligner loss after 1,500 epochs (roughly after 60\%) of the training, and continued to train with other loss functions.
We observed that using this training schedule improved the quality for Hindi while not affecting the results for Tamil, and hence we use this for all subsequent models.



\subsection{Evaluation of single speaker and multi-speaker models}
\label{subsec:speaker}
We train multi-speaker models with one male and one female voice with speaker embeddings jointly learnt and added to the encoder's output.
As can be seen in Table~\ref{tab:speaker}, multi-speaker models have better scores for both languages and both speakers.
This suggests the value of joint training and efficiently deploying models for both male and female speakers.
We hypothesize that the improved performance is because the aligner module not being conditioned on the speaker embeddings learns better alignments on the more diverse data of both genders.

\begin{table}[!htbp]
\centering
\begingroup
\setlength{\tabcolsep}{6pt} 
\renewcommand{\arraystretch}{0.9} 
\begin{tabular}{@{}lcccc@{}}
\toprule
\textbf{Language}  & \multicolumn{2}{c}{\textbf{Female}} & \multicolumn{2}{c}{\textbf{Male}} \\ 
 \cmidrule(l){2-3} \cmidrule(l){4-5}
                   & Single    & Multi   & Single   & Multi  \\ \midrule
Tamil (Dravidian)  & 3.55              & 3.71            & 3.84             & 3.98           \\ 
Hindi (Indo-Aryan) & 3.73              & 4.02            & 3.82             & 3.98           \\
\bottomrule
\end{tabular}
\endgroup
\caption{Subjective evaluation (MOS) of our single-speaker and multi-speaker models for Tamil and Hindi.}
\label{tab:speaker}
\end{table}

\subsection{Evaluation of monolingual and multilingual models}
\label{subsec:multilingual}
In Table \ref{tab:multilingual}, we report the results of the subjective evaluation of multilingual models for the two groups w.r.t. monolingual models. We find it encouraging that multilingual models achieve similar MOS in Kannada, Tamil, Telugu, and Assamese.
However, overall the monolingual models outperform in all languages except Gujarati.
We thus choose to train monolingual models.


\begin{table}[htbp]
\centering
\begingroup
\setlength{\tabcolsep}{6pt} 
\renewcommand{\arraystretch}{0.9} 
\begin{tabular}{lcccr}
\toprule
\textbf{Language} & \textbf{Mono} & \textbf{Multi} & \textbf{Multi} & \textbf{Gap} \\
&  & (Dravidian) & (Indo-Aryan) & \\ \midrule
Kannada   & 3.68 & 3.60  & -    & 0.08  \\
Malayalam & 3.64 & 3.09 & -    & 0.55  \\
Tamil     & 3.84 & 3.80  & -    & 0.04  \\
Telugu    & 3.66 & 3.56 & -    & 0.10   \\ \midrule
Assamese  & 2.39 & -    & 2.33 & 0.06  \\

Bengali   & 3.37 & -    & 2.95 & 0.42  \\
Gujarati  & 3.58 & -    & 3.66 & -0.08 \\
Hindi     & 4.00    & -    & 3.19 & 0.81  \\
Marathi   & 3.26 & -    & 2.62 & 0.64  \\
Odia      & 4.19 & -    & 3.5  & 0.69   \\ \bottomrule
\end{tabular}
\endgroup
\caption{Subjective evaluation (MOS) of our monolingual models and multilingual models.}
\label{tab:multilingual}
\end{table}

\subsection{Comparison of open-source Indic TTS models}
\label{subsec:best}
Finally, based on our findings, for each of the 13 Indian languages, we train monolingual multi-speaker models with FastPitch and HiFiGAN V1 \footnote{\scriptsize \url{https://models.ai4bharat.org/\#/tts/samples}}, with no supplementary losses, but with the aligner loss turned off after 1500 epochs. 
In Table \ref{tab:main}, we compare our model against existing open-source TTS models trained on the IndicTTS Dataset.
We see that our model is clearly rated better with an average MOS score improvement of 0.51 w.r.t. models proposed in \cite{prakash2020generic} and 0.92 w.r.t. models proposed in \cite{vakyansh2021glow}.
The objective metrics of $F_0$ and CER also follow a similar trend. 
The MCD scores however show a lower value for models from \cite{vakyansh2021glow}, but this is uncorrelated to MOS scores \cite{salesky2021assessing} and the audio samples are indeed unnatural.

\section{Conclusion}
Neural TTS systems continue to rapidly improve with various changes. 
We evaluated the choice of acoustic models, vocoders, training strategies, and multi-speaker and multilingual generalizations for Indian languages. 
With the identified best configuration we train models for 13 Indian languages for both genders and establish that it improves on existing TTS systems.
We open-source the models for the 13 languages on the Bhashini platform enabling applications targeting over 1.05 billion native speakers as per 2011 census.

Several directions of future work emerge.
Diffusion-based acoustic models and flow-based vocoders need to be compared. 
Further exploration is required for sharing knowledge while training multilingual TTS models which have a clear advantage in deployability. 
Open-source models for expressive speech, voice cloning, and unheard speaker generalization for Indian languages remain to be thoroughly investigated.

\section{Acknowledgements}

We would like to thank the Ministry of Electronics and Information Technology (MeitY\footnote{\scriptsize \url{https://www.meity.gov.in/}}) of the Government of India and the Centre for Development of Advanced Computing (C-DAC\footnote{\scriptsize \url{https://www.cdac.in/index.aspx?id=pune}}), Pune for generously supporting this work and providing us access to multiple GPU nodes on the Param Siddhi Supercomputer. We would like to thank the EkStep Foundation and Nilekani Philanthropies for their generous grant which went into hiring human resources as well as cloud resources needed for this work. We would like to thank Janki Nawale from AI4Bharat for helping in coordinating the evaluation tasks and extend our gratitude to all the language experts of the AI4Bharat team who helped in the evaluation.


\bibliographystyle{IEEEbib}
\bibliography{refs}

\newpage\phantom{dummy}

\begin{figure*}[!h]
    \centering
    \includegraphics[width=0.8\textwidth]{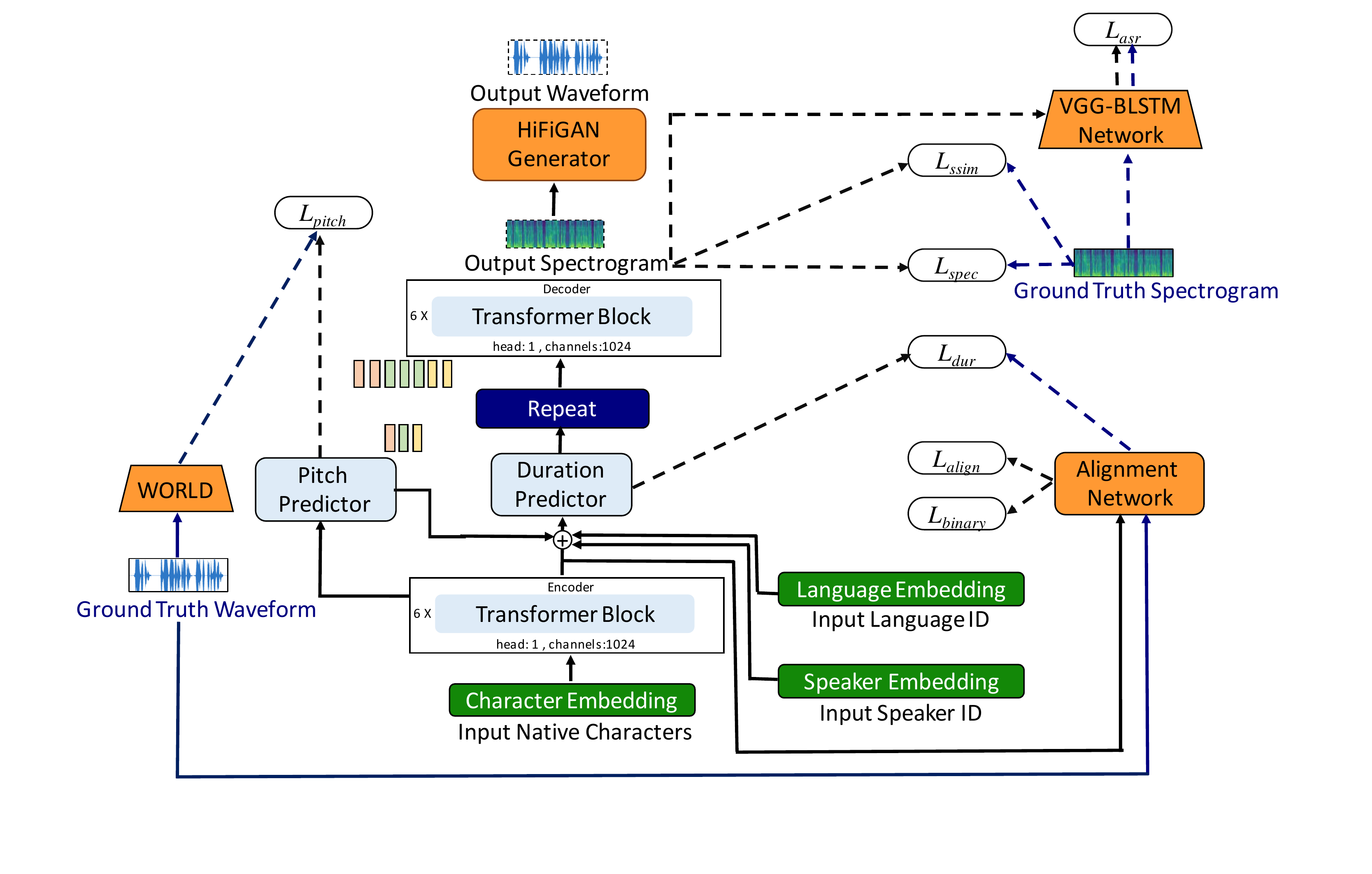}
    \caption{Unified architecture of our TTS system.}
    \label{fig:unified}
\end{figure*}

\newpage
\appendix
\section{Appendix}
\subsection{Unified Architecture}
\label{appendix:unified}

In Figure \ref{fig:unified}, our final model is depicted, comprising of FastPitch \cite{lancucki2021fastpitch} as the acoustic model, the alignment learning framework \cite{badlani2022one} to map text features to mel-spectrograms and HiFiGAN-V1 \cite{kong2020hifi} as the vocoder.

 The encoder maps the input text into a hidden representation $\mathbf{h}$, which is used by the pitch predictor and duration predictor to infer the average pitch and duration for every input symbol respectively. The encoder consists of 6 Transformer blocks each with one head, hidden channel dimension of 1024, and a dropout of 0.1. The pitch predictor and the duration predictor are trained using mean-squared-error losses, $L_{pitch}$ and $L_{dur}$ respectively. The pitch information is added to $\mathbf{h}$ and the resulting vector is discretely upsampled using the duration information. The decoder then operates on the upsampled vector and outputs the generated mel-spectrogram $\Hat{\mathbf{m}}$ which is reconstructed, using ground-truth mel-spectrograms $\mathbf{m}$ as a reference, by the reconstruction loss $L_{spec}$ given by,
\[
L_{spec} = \left \lVert \mathbf{m} - \Hat{\mathbf{m}} \right \rVert_{2}^{2}
\]

 Unlike the original FastPitch \cite{lancucki2021fastpitch}, which relies on an external aligner such as a pretrained Tacotron2, we employ an alignment learning framework \cite{badlani2022one} to align encoded text features $\mathbf{h}$ to mel-spectrograms $\mathbf{m}$. To do this, a soft alignment distribution $\mathcal{A}_{soft}$
based on the learned pairwise affinity between encoded hidden representation 
and encoded mel-spectrogram frames is first computed and it is normalized using softmax across the domain of input text. An objective function is then used that maximizes the likelihood of the encoded hidden representation given mel-spectrograms using the forward-sum algorithm and its negative is defined as the aligner loss $L_{aligner}$. We also use the Viterbi algorithm as in \cite{badlani2022one}, to convert soft alignments  to hard alignments $\mathcal{A}_{hard}$ and use an additional binarization loss $L_{binary}$ that minimizes the KL-Divergence between $\mathcal{A}_{soft}$ and $\mathcal{A}_{hard}$. The aligner is not conditioned on speaker embeddings or language embeddings and takes text embeddings from the text encoder.
\[
L_{binary} = - \mathcal{A}_{hard} \odot log \mathcal{A}_{soft}
\]

Let $L_{ssim}$ refer to the perceptual SSIM loss briefly discussed in Section \ref{subsec:training-strategies}. The ASR loss $L_{asr}$ minimizes the manhattan distance between convolutional features $F_{asr}$ of the ground truth mel-spectrograms $\mathbf{m}$ and synthesized mel-spectrograms $\Hat{\mathbf{m}}$,
extracted from the intermediate layer before the LSTM layers of a pretrained joint CTC-attention VGG-BLSTM network provided in the Espnet toolkit.
\[
L_{asr} = \left \lVert F_{asr}(\mathbf{m}) - F_{asr}(\Hat{\mathbf{m}}) \right \rVert_{1}
\]
The final loss $L_{acoustic}$ is given by,
\begin{align*}
    L_{acoustic} &=  L_{spec} + \lambda_{align}L_{align} + \lambda_{dur}L_{dur} + \lambda_{pitch}L_{pitch} &\\
    & + \lambda_{binary}L_{binary} + \lambda_{ssim}L_{ssim} + \lambda_{asr}L_{asr} &
\end{align*}
We set, $\lambda_{dur}=0.1$, $\lambda_{pitch}=0.1$, $\lambda_{binary}=0.1$. For experiments in Section \ref{subsec:loss} we set $\lambda_{ssim}=1$ and $\lambda_{asr}=0.5$.

We train our acoustic model and vocoder separately and put them in sequence for inference. We use HiFi-GAN V1 \cite{kong2020hifi} as a vocoder to generate high-fidelity speech from mel-spectrograms. For HiFi-GAN V1 training, we use the ground truth spectrogram as input and learn the generator to produce the ground truth waveform. We train HiFi-GAN with the same objective functions for the generator and discriminators as the original work \cite{kong2020hifi}, where the generator objective function is the combination of GAN loss, feature matching loss and the mel-spectrogram loss. Although the vocoder is not conditioned on either speaker or language embeddings, it is able to generalize well.


\end{document}